\documentclass{article}

\usepackage{arxiv}

\usepackage{color,array}
\usepackage{hyperref}
\usepackage{colortbl} 
\usepackage{graphicx}
\usepackage{amssymb}
 \usepackage{multirow}
\usepackage{lipsum}
\usepackage{mathtools}
\usepackage{microtype}
\usepackage{graphicx}
\usepackage{hyperref}
\usepackage{subfigure}
\usepackage{booktabs}
\usepackage{longtable}
\usepackage{algorithmic}
\usepackage{algorithm}
\usepackage{dirtytalk}
\usepackage{xcolor}
\usepackage{siunitx}
\usepackage{caption}
\usepackage{float}
\usepackage{adjustbox}
\usepackage{placeins}
\usepackage{balance}
\usepackage{textcomp}
\newtheorem{theorem}{Theorem}[section]

\title{Adapting to Fragmented and Evolving Data: A Fisher Information Perspective }

\author{
    \href{https://orcid.org/0000-0003-0985-9543}{\includegraphics[scale=0.06]{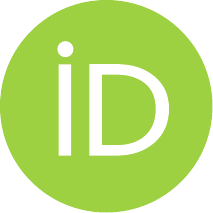}\hspace{1mm}Behraj Khan}\thanks{All authors contributed equally to this work.},  
    \href{https://orcid.org/0000-0003-0638-9689}{\includegraphics[scale=0.06]{orcid.pdf}\hspace{1mm}Tahir Syed} \\
    School of Mathematics and Computer Science, Institute of Business Administration Karachi, Pakistan \\
    \texttt{\{behrajkhan, tahirqsyed\}@gmail.com} \\
    \href{https://orcid.org/0000-0001-6135-3924}{\includegraphics[scale=0.06]{orcid.pdf}\hspace{1mm}Nouman Durrani} \\
   National University of Computer and Emerging Sciences, Karachi \\
    \texttt{behroz.mirza@sse.habib.edu.pk}
}

\renewcommand{\shorttitle}{\textbf{FADE:} \textit{Fisher-based Adaptation to Dynamic Environments}}

\hypersetup{
pdftitle={A template for the arxiv style},
}

\begin{document}
\maketitle

\begin{abstract}

Modern machine learning systems deployed in dynamic environments face \emph{Sequential Covariate Shift (SCS)}, where input distributions evolve over time. We propose \textbf{FADE} (\textit{Fisher-based Adaptation to Dynamic Environments}), a lightweight and theoretically grounded framework for robust learning under such conditions. FADE introduces a shift-aware regularization scheme anchored in Fisher information geometry, guiding adaptation by weighting parameter updates according to sensitivity and stability. To detect meaningful distribution changes, we derive a Cramér-Rao-informed shift signal combining KL divergence with temporal Fisher dynamics. Unlike prior methods that require task boundaries, target supervision, or replay, FADE operates online with fixed memory and no target labels. Evaluated on seven benchmarks across vision, language, and tabular modalities, FADE achieves up to 19\% higher accuracy under severe shifts, consistently outperforming state-of-the-art baselines like TENT and DIW. FADE further extends naturally to federated learning by treating heterogeneous clients as temporally fragmented environments, delivering scalable and stable adaptation in decentralized settings. Theoretical analysis guarantees bounded regret and parameter consistency, while empirical results demonstrate FADE’s robustness across modalities and shift intensities.

\end{abstract}


\maketitle
\section{Introduction}
\label{sec:introduction}

Many real-world machine learning (ML) applications such as federated learning, edge computing, and time-series forecasting encounter fragmented data that arrives in sequential batches. This gives rise to \textit{Sequential Covariate Shift (SCS)}, where the input distribution \( P(x) \) changes over time while the conditional distribution \( P(y|x) \) remains fixed \cite{shimodaira2000improving,sugiyama2007covariate}. SCS violates the i.i.d.\ assumption and can lead to substantial performance degradation. Addressing SCS is crucial for ensuring robust generalization in dynamic environments \cite{khan2024causal}.

In federated learning, for example, clients may represent geographically diverse or demographically distinct populations, resulting in non-identically distributed local data. These distributional biases are often ignored in traditional ML pipelines that assume static, centralized data \cite{gupta2022fl}. Our work examines the consequences of such distributional disparity and proposes techniques to mitigate its impact, aiming for consistent predictive performance across evolving subsets of data.

Standard model selection techniques like cross-validation are reliable under i.i.d.\ assumptions \cite{bishop1995neural}, but they break down in realistic settings where distributional shift is common \cite{van2015fast}. We focus on scenarios where the training data undergoes multiple covariate shifts as in federated and online learning while the validation/test distribution remains stationary. This formulation aligns with classical \textit{covariate shift} problems but introduces the added complexity of sequential, non-stationary training batches.

Covariate shift, also known as \textit{sample selection bias} \cite{cortes2008sample}, \textit{population shift} \cite{kelly1999impact, hand2006classifier}, or \textit{non-stationary data} \cite{cieslak2009framework}, refers to the setting where conditional distribution remains same \( P_{tr}(y|x) = P_{tst}(y|x) \) but feature distribution changes between train and test time \( P_{tr}(x) \neq P_{tst}(x) \) \cite{moreno2012unifying, quinonero2009dataset}. This shift is common in domains like emotion recognition \cite{jirayucharoensak2014eeg}, speech processing \cite{yang2007weighted}, BCI \cite{li2010application}, and spam detection \cite{bickel2006dirichlet}.

While many studies have addressed covariate shift detection in offline \cite{rabanser2019failing, hu2020distribution, moreno2012study} and online \cite{vovk2020testing, vovk2021testing} settings, fewer have tackled the unique challenges of sequential shift during training. Sugiyama et al.\ \cite{sugiyama2007covariate, sugiyama2012density} proposed importance weighting to address shift in cross-validation, but such methods require access to the test distribution and may be brittle under dynamic scenarios.

Efforts to combat client heterogeneity in federated learning via techniques like gradient variance reduction \cite{karimireddy2020scaffold} or knowledge distillation \cite{li2020federated} often falter when facing SCS. Similarly, large-scale systems like Lambda architecture \cite{marz2013big}, Spark \cite{zaharia2010spark}, and Apache Mahout \cite{owen2011mahout} focus on computation but overlook statistical shift. As streaming and distributed systems scale, fragmented and evolving datasets become the norm, making SCS a pressing ML challenge.

Traditional ML pipelines assume consistent training and test distributions. However, real-world data collected from edge devices or evolving time-indexed streams violates this assumption, causing biased estimates and poor generalization \cite{gupta2022fl}. SCS naturally arises when:
\begin{itemize}
    \item Data is collected over time (e.g., streaming/online learning)
    \item Federated clients hold non-i.i.d.\ local datasets
    \item Underlying populations evolve during deployment
\end{itemize}

\begin{figure}[htbp]
\centering
\includegraphics[width=0.5\textwidth]{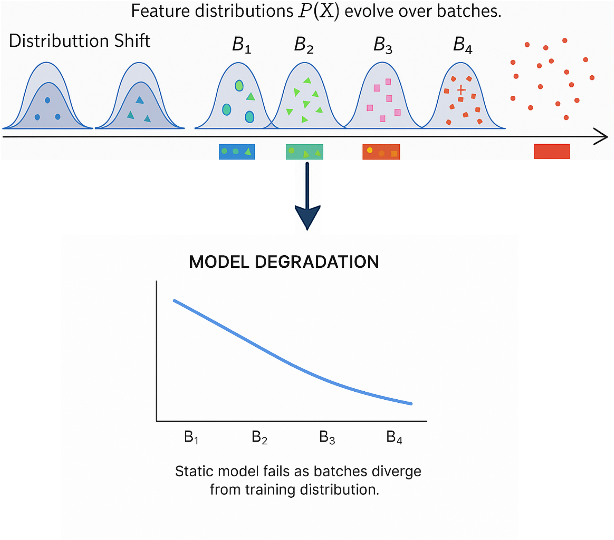}
\caption{Sequential Covariate Shift (SCS): Model performance drops as input distributions evolve over training batches.}
\label{fig:detection}
\end{figure}

Classical solutions like cross-validation or importance weighting \cite{shimodaira2000improving} assume access to i.i.d.\ samples or target distributions, assumptions that rarely hold in sequential or decentralized settings. Figure~\ref{fig:detection} illustrates how model performance drops across training batches as input distributions shift.

\begin{figure*}[htbp]
\centering
\includegraphics[width=0.6\textwidth]{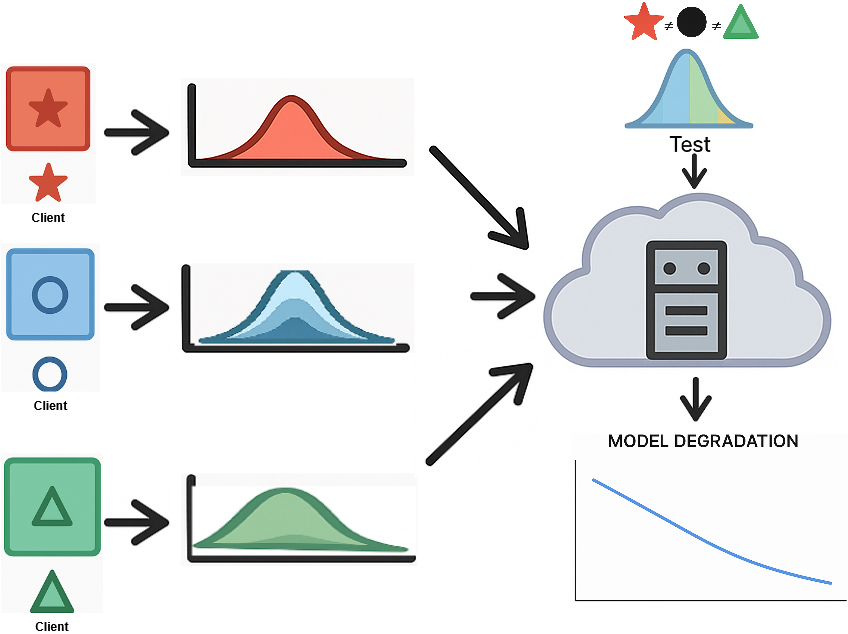}
\caption{Client-wise heterogeneity in federated learning causes model degradation. Each symbol denotes a different client's data; shifts among them highlight SCS (\textcolor{red}{$\star$} $\neq$ \textcolor{blue}{$\bigcirc$} $\neq$ \textcolor{green}{$\triangle$}).}
\label{fig:image1}
\end{figure*}

In federated learning, clients often represent distinct user profiles. As shown in Figure~\ref{fig:image1}, each client (\textcolor{red}{$\star$}, \textcolor{blue}{$\bigcirc$}, \textcolor{green}{$\triangle$}) generates data from different distributions, inducing server-side model degradation due to heterogeneity.

Existing solutions like domain adaptation \cite{ganin2016domain} or gradient correction \cite{karimireddy2020scaffold} assume centralized data or synchronized updates. Systems like Spark \cite{zaharia2010spark} or Mahout \cite{owen2011mahout} enhance efficiency but overlook dynamic covariate shifts.

Our work tackles a more realistic and challenging setting: training over voluminous, fragmented datasets with unknown and evolving covariate shifts. Specifically, we address the problem of detecting and adapting to SCS when data is assimilated in a fixed, sequential batch order.

We introduce a novel perspective by linking covariate shift with the Cramér-Rao lower bound. We observe that the Fisher Information Matrix (FIM), which bounds the variance of unbiased estimators, also encodes model sensitivity to distributional changes. This insight motivates our central contribution: a lightweight regularization framework called \textbf{FADE} (Fisher-based Adaptation to Dynamic Environments) that:

\begin{itemize}
    \item Detects distributional shifts using KL divergence weighted by FIM changes
    \item Adapts parameters through batch-wise, FIM-informed optimization
    \item Operates with fixed memory and computational cost per batch
\end{itemize}

FADE is model-agnostic, requires no tuning for shift timing or magnitude, and integrates easily into standard training pipelines. We evaluate FADE on diverse domains including image classification, text (Amazon reviews), and tabular benchmarks from OpenML\footnote{\url{https://www.openml.org/}}. Experiments show consistent improvements over baselines, with up to 19\% accuracy gains under high-shift scenarios.

\subsection*{Contributions}
\begin{enumerate}
\item We introduce FADE, a novel regularization-based method that uses Fisher Information and KL divergence to adapt to sequential covariate shifts, achieving up to 19\% higher accuracy than state-of-the-art approaches.

\item We formally define and analyze Sequential Covariate Shift (SCS) caused by dataset fragmentation, providing both theoretical justification and empirical validation of its impact on model performance.

\item Our method efficiently processes fragmented datasets in batches with linear memory complexity, requiring only two batches in memory at any time while maintaining robust performance.

\item We demonstrate through extensive experiments on 7 benchmark datasets that FADE effectively handles both natural covariate shifts and those induced by data fragmentation, with consistent improvements across different batch sizes.

\item We provide complete implementation code and experimental protocols to ensure reproducibility and facilitate adoption in real-world applications.
\end{enumerate}

The rest of the paper is organized as follows: Section 2 reviews related work. Section 3 formulates SCS and our approach. Section 4 details the FADE algorithm. Sections 5 and 6 present experimental results and analysis. Section 7 concludes with limitations and future directions.

\section{Sequential Covariate Shift: Detection and Adaptation}
\label{sec:scs}
\subsection{Problem Characterization}
Sequential Covariate Shift (SCS) arises in incremental learning settings where data arrives as temporally ordered batches $\{\mathcal{D}_1, \mathcal{D}_2, \ldots, \mathcal{D}_T\}$, each with potentially distinct feature distributions. We formally define SCS at time $t$ as:

\begin{equation}
\Delta_{SCS}^{(t)} = D_{KL}(P_t(X) \| P_{t+1}(X)) > \epsilon,
\end{equation}

where $P_t(X)$ denotes the feature distribution of batch $\mathcal{D}_t$, $D_{KL}$ is the Kullback-Leibler divergence, and $\epsilon$ is a shift significance threshold. SCS differs from classical covariate shift in three key aspects:

\begin{itemize}
    \item \textbf{Order Dependency:} Batches arrive in an immutable sequence, inducing path-dependent model updates.
    \item \textbf{Partial Observability:} At time $t$, only batches $\{\mathcal{D}_1, \ldots, \mathcal{D}_t\}$ are accessible.
    \item \textbf{Compounding Shift:} Small shifts accumulate over time, potentially leading to large distributional shift.
\end{itemize}

\subsection{Detection Framework}
To detect SCS, we leverage the Fisher Information Matrix (FIM), which quantifies local sensitivity of the model likelihood to parameter changes:

\begin{equation}
I_t(\theta) = -\mathbb{E}_{x \sim P_t} \left[ \nabla_\theta^2 \log p_\theta(x) \right].
\end{equation}

We define a shift signal $\tau_t$ that combines distributional and geometric changes:

\begin{equation}
\tau_t = \|I_t(\theta) - I_{t-1}(\theta)\|_F \cdot D_{KL}(P_t \| P_{t-1}),
\end{equation}

where $\|\cdot\|_F$ is the Frobenius norm. A shift is flagged if $\tau_t > \gamma$, where $\gamma$ is a predefined threshold. This hybrid criterion stabilizes detection in high dimensions, where KL divergence alone may be insufficient. Figure \ref{fig:SCS_pipeline} illustrates the pipeline for detecting  SCS using a combined signal approach. Initially, the divergence between training and test feature distributions is quantified using KL divergence, where sudden spikes (Figure a) indicate potential distributional shift. To complement this, the FIM is computed across timestep or batches, capturing structural changes in model sensitivity to input features (Figure b). By integrating these two indicators distributional deviation and model-aware sensitivity into a unified detection metric (Figure c), the pipeline provides a more robust and timely signal for identifying harmful shifts. The final detection curve surpasses a calibrated threshold (\textcolor{red}{red dashed line}), enabling automatic triggering of adaptation mechanisms. This composite strategy enhances both the sensitivity and specificity of shift detection in evolving data streams.

\begin{figure}[htbp]
\centering
\includegraphics[width=0.5\textwidth]{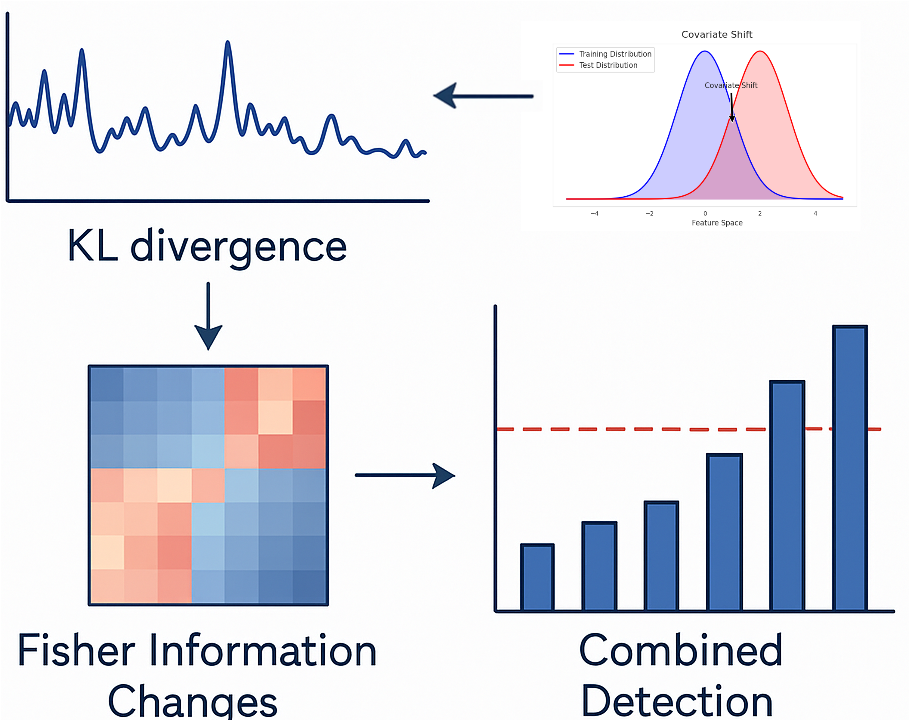}
\caption{SCS detection pipeline: (a) KL divergence spikes represent potential distribution shift, (b) FIM variations captures changes in model sensitivity of input features, (c) combined detection signal for triggering automatic adaptation mechanism.}
\label{fig:shiftdetection}
\end{figure}

\subsection{Adaptation via Information Preservation}
Adaptation is guided by the Cramér-Rao bound, which states that the covariance of any unbiased estimator $\hat{\theta}$ satisfies:

\begin{equation}
\text{Cov}(\hat{\theta}) \succeq I(\theta)^{-1}.
\end{equation}

We maintain a moving estimate of a global FIM via exponential smoothing:

\begin{equation}
I_{global}^{(t)} = \alpha I_{global}^{(t-1)} + (1 - \alpha) I_t(\theta),
\end{equation}

where $\alpha \in [0,1]$ controls temporal sensitivity. Using this, we regularize the objective at time $t$ as:

\begin{equation}
\mathcal{L}_t(\theta) = \underbrace{\mathbb{E}_{(x,y) \sim P_t}[\ell(f_\theta(x), y)]}_{\text{Current task loss}} + \lambda \underbrace{(\theta - \mu_{t-1})^\top I_{global}^{(t-1)} (\theta - \mu_{t-1})}_{\text{CRB regularizer}},
\end{equation}

where $\mu_{t-1}$ denotes the previous parameter estimate and $\lambda$ is a regularization strength.

\subsection{Theoretical Guarantees}
The FADE framework offers the following theoretical properties under mild assumptions:

\begin{theorem}[Bounded Regret]
Assuming Lipschitz-continuous loss functions, the cumulative regret after $T$ batches satisfies:
\begin{equation}
R(T) \leq \sqrt{T}(1 + \log T) + \frac{1}{\lambda} \sum_{t=1}^T \Delta_{SCS}^{(t)}.
\end{equation}
\end{theorem}

\begin{theorem}[Parameter Consistency]
As $T \rightarrow \infty$, the model parameters converge to the optimal solution for the global distribution:
\begin{equation}
\|\theta_T - \theta^*\|_2 = \mathcal{O}(1/\sqrt{T}).
\end{equation}
\end{theorem}

\subsection{Computational Considerations}
FADE maintains scalability via:

\begin{itemize}
    \item Diagonal approximation of the FIM for deep models
    \item Layer-wise selective updates based on shift detection
    \item Warm-starting optimization using previous batch parameters
\end{itemize}

These choices ensure per-batch complexity is $\mathcal{O}(d)$, where $d$ is the number of parameters, enabling application to large-scale models.

\begin{algorithm}[t]
\caption{FADE: Sequential Adaptation}
\begin{algorithmic}[1]
\REQUIRE Data stream $\{\mathcal{D}_1, \ldots, \mathcal{D}_T\}$, initial parameters $\theta_0$
\STATE Initialize $I_{global} \leftarrow \epsilon I_d$ (small diagonal matrix)
\FOR{$t = 1$ to $T$}
    \STATE Compute $I_t(\theta_{t-1})$ on $\mathcal{D}_t$ (diagonal approx.)
    \STATE Detect shift: $\tau_t = \|I_t - I_{t-1}\|_F \cdot D_{KL}(P_t \| P_{t-1})$
    \IF{$\tau_t > \gamma$}
        \STATE Update $I_{global} \leftarrow \alpha I_{global} + (1 - \alpha) I_t$
        \STATE Optimize: $\theta_t \leftarrow \operatorname*{arg\,min}_\theta \mathcal{L}_t(\theta)$
    \ELSE
        \STATE $\theta_t \leftarrow \theta_{t-1}$ \hfill 
    \ENDIF
\ENDFOR
\end{algorithmic}
\end{algorithm}

\subsection{Connection to Federated Learning}
Federated Learning (FL) involves decentralized training across multiple clients, each with private, potentially non-identically distributed (non i.i.d.) data \cite{mcmahan2017communication}. A central challenge in FL is managing \emph{statistical heterogeneity}-distributional differences across client data-which degrades convergence and generalization. Our proposed framework, while not developed specifically for FL, naturally extends to this setting by treating each client's local data as a sequential batch in the SCS framework.

Let $\mathcal{D}_t$ represent the local dataset from client $t$, and assume local distributions $P_t(X)$ vary across clients due to demographic, temporal, or usage-related factors. The global objective in FL is to minimize the aggregated empirical risk:

\begin{equation}
\min_{\theta} \sum_{t=1}^T \frac{n_t}{n} \mathbb{E}_{(x,y) \sim P_t}[\ell(f_\theta(x), y)],
\end{equation}

where $n_t = |\mathcal{D}_t|$ and $n = \sum_t n_t$. Traditional FL methods such as FedAvg \cite{mcmahan2017communication} assume synchronized updates and often underperform in the presence of sharp client-wise distribution shifts.

In contrast, FADE provides a principled mechanism to address this heterogeneity by modeling the shift between successive clients via the Fisher-weighted KL divergence:

\begin{equation}
\tau_t = \|I_t(\theta) - I_{t-1}(\theta)\|_F \cdot D_{KL}(P_t \| P_{t-1}),
\end{equation}

which now captures client-to-client distribution variation. When $\tau_t > \gamma$, the model adapts via a Fisher-regularized objective that preserves parameter stability across heterogeneous clients. This can be interpreted as introducing a form of client-wise information anchoring, mitigating catastrophic forgetting during federated aggregation.

Moreover, because FADE operates without access to future distributions and with constant memory, it aligns well with the privacy and resource constraints in FL. Each client can apply local adaptation with only their own data, and the server can aggregate model updates without needing direct access to raw data or client-specific distributions.

\begin{figure*}[htbp]

        \centering
        \includegraphics[width=0.8\textwidth]{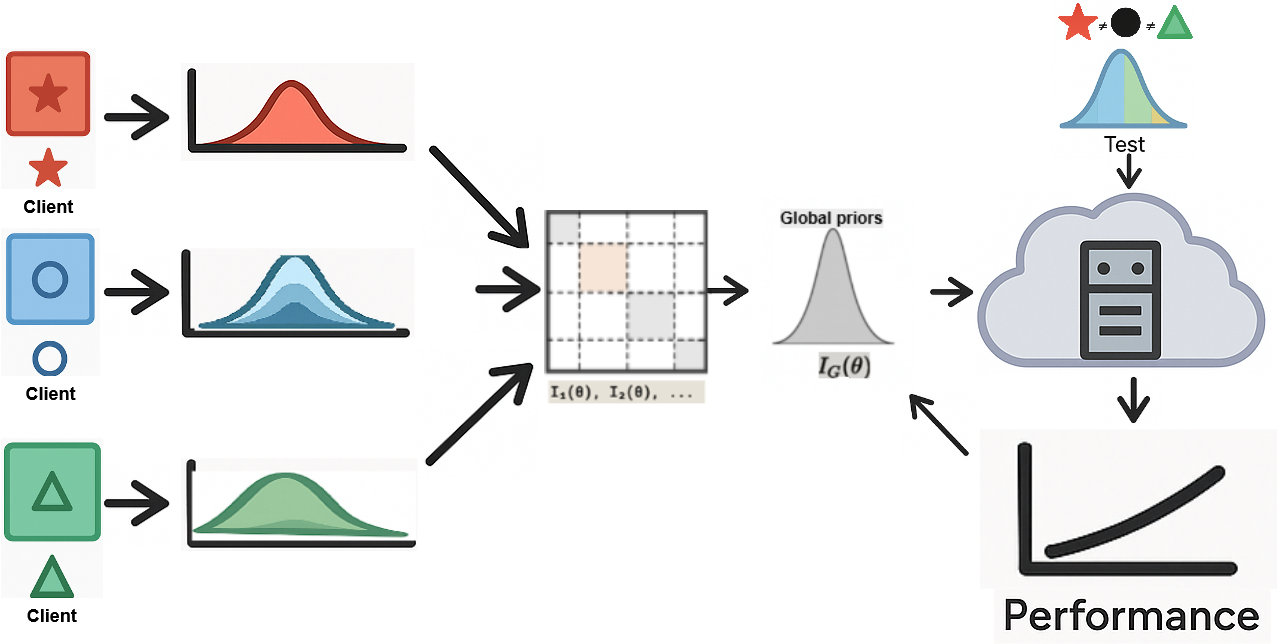}
 \caption{Workflow of the proposed FADE framework: FADE extends to Federated Learning by modeling client-wise such as \textcolor{red}{\small $\star$}  represent client 1,  \textcolor{blue}{\small $\bigcirc$}  client 2 and  \textcolor{green}{\small $\triangle$} represent client 3 data as sequential shifts (as it can be noticed that test set distribution different from training (\textcolor{red}{\small $\star$} $\neq$ \textcolor{blue}{\small $\bigcirc$} $\neq$ \textcolor{green}{\small $\triangle$}). FADE uses FIM (\(I(\theta)\)) to adapt while preserving stability, without violating privacy constraints.}

    \label{fig:combined}
\end{figure*}

Thus, FADE provides a theoretically grounded and practically efficient strategy for adapting to distributional heterogeneity in federated learning, bridging sequential shift modeling with decentralized optimization. The working mechanism of our framework is given in figure \ref{fig:combined}.

\section{Related Work}

\label{sec:related}

FADE connects the domains of covariate shift adaptation, sequential learning, and distributed optimization particularly within federated settings. While prior work has addressed components of this problem space, existing methods fall short of providing a unified solution for online covariate shift adaptation with both theoretical guarantees and practical scalability. We contextualize FADE by comparing it with prior work across four primary areas.

\textbf{Tabular Learning under Covariate Shift.} Traditional methods for covariate shift often rely on access to both source and target distributions to estimate importance weights. Empirical Risk Minimization (ERM)\cite{vapnik1998statistical}, while foundational, is agnostic to distribution shift and thus leads to suboptimal generalization under shift. Kernel-based approaches such as uLSIF and RuLSIF\cite{yamada2013relative,sugiyama2012density} offer theoretical guarantees for density ratio estimation but are computationally impractical in high-dimensional settings. Deep variants like One-step~\cite{fang2020rethinking} perform dynamic reweighting for adaptation but require batch access to the target domain, rendering them unsuitable for streaming scenarios. FADE avoids explicit density estimation and enables online, target-free adaptation across evolving distributions, generalizing these classical methods.

\textbf{Online Adaptation in Vision.} In the vision domain, many adaptation strategies operate at test time using feature realignment or entropy minimization. ERM again serves as a baseline but lacks post-deployment adaptability. Elastic Weight Consolidation (EWC)\cite{kirkpatrick2017overcoming} uses the Fisher Information Matrix (FIM) to regularize updates and prevent forgetting, though it assumes discrete task boundaries and is not designed for gradual or continuous shifts. Test-time Entropy Minimization (TENT)\cite{wang2020tent} modifies batch normalization layers to adapt online, yet it lacks memory and formal guarantees, and often suffers from instability. DIW~\cite{fang2020rethinking} improves robustness through sample reweighting but assumes access to offline target data. FADE overcomes these limitations by combining Fisher-based online shift detection with a regularized update strategy that avoids reliance on task segmentation or explicit memory.

\textbf{Federated Learning and Distributed Shifts.} Federated learning introduces unique challenges when clients experience non-iid covariate shifts. FedAvg~\cite{mcmahan2017communication} aggregates client updates naively, often resulting in subpar convergence under heterogeneity. FedProx~\cite{li2020federated} introduces proximal regularization for stability, yet does not model distributional shifts explicitly. SCAFFOLD~\cite{karimireddy2020scaffold} addresses client shift using control variates but similarly assumes static distributions. In contrast, FADE incorporates client-specific covariate shift detection and adaptation via Fisher-weighted updates, achieving effective personalization without requiring target data sharing or centralized coordination.

\textbf{Theoretical Foundations.} FADE draws inspiration from online learning theory, particularly regret minimization frameworks~\cite{zinkevich2003online}, as well as from information geometry, leveraging Fisher Information as a sensitivity metric~\cite{amari2016information}. It also builds upon continual learning principles, notably EWC~\cite{kirkpatrick2017overcoming}, while addressing its limitations such as the need for predefined task boundaries. FADE unifies these theoretical components to yield a streaming algorithm with formal guarantees for both shift detection and adaptation in high-dimensional, decentralized environments.

\textbf{Distinguishing Features and Advantages.} Unlike batch-based reweighting methods like DIW and uLSIF, FADE processes data sequentially and scales efficiently to high-dimensional domains. It avoids computational burdens associated with kernel density estimation and adversarial training (e.g., GAN-based DA~\cite{ganin2016domain}). In federated settings, it explicitly models client-specific shifts a capability absent in popular methods like FedAvg or SCAFFOLD. Furthermore, it provides formal guarantees for both adaptation and shift detection, positioning it as a theoretically grounded and practically effective solution for real-world continual learning under distribution shift.

\section{Experimental Setup}
\label{sec:setup}


\subsection{Datasets}
We evaluate FADE across seven datasets spanning vision, language, and tabular modalities, selected for their diverse covariate shift properties and relevance to dynamic learning scenarios. Each dataset is partitioned into temporally ordered batches $\{\mathcal{D}_1, \ldots, \mathcal{D}_T\}$ to simulate sequential non-i.i.d. data arrival.

\textbf{Vision. }
We use three benchmarks: (1) MNIST~\cite{lecun1998mnist}, with synthetic covariate shifts induced via rotations and additive noise; (2) MNIST-M~\cite{ganin2016domain}, a domain-shifted variant of MNIST with colored backgrounds; and (3) CIFAR-10/100~\cite{krizhevsky2009learning}, partitioned into class-incremental batches to emulate gradual semantic shift.

\textbf{Language. }
We use the Amazon Reviews dataset~\cite{hou2024bridging}, containing sentiment-labeled product reviews. We introduce sequential semantic shifts by splitting data across product categories such as books, electronics, and apparel.

\textbf{Tabular. }
We include: (1) Credit Card Fraud~\cite{yadav29credit}, a highly imbalanced transaction dataset with natural temporal ordering; and (2) five classic datasets from \textit{OpenML}~\cite{bischl2017openml} (e.g., australian,breast cancer, diabetes, heart, sonar ) modified with feature masking and class imbalance to simulate structured shifts.

\textbf{Federated Learning. }
For distributed experiments, we extend evaluation to: (1) FEMNIST~\cite{caldas2018leaf}, a handwriting dataset partitioned by user; (2) SVHN~\cite{netzer2011reading}, with client splits based on digit identities; and (3) Shakespeare~\cite{mcmahan2017communication}, a character-level language modeling dataset with client-wise partitioning by speaker.

\textbf{Shift Characterization. }
We quantify covariate shift intensity using batch-wise KL divergence $\Delta_{\mathrm{SCS}}^{(t)}$, and organize evaluation scenarios into mild, moderate, and severe regimes (see Table~\ref{tab:severity}).

\begin{table*}[!ht]
\centering
\caption{Benchmarking with SOTA on tabular datasets under covariate shift. Mean accuracy (\%) with standard deviation. Bold indicates best performance.}
\label{tab:performance}
\begin{tabular}{lccccc}
\hline
\textbf{Dataset} & \textbf{ERM} & \textbf{uLSIF} & \textbf{RuLSIF} & \textbf{One-step} & \textbf{FADE (Ours)}  \\
\hline
australian & 67.9 ± 16.8 & 69.3 ± 16.3 & 69.6 ± 15.1 & 74.4 ± 12.7 & \textbf{75.1 ± 5.92}  \\
breast cancer & 78.3 ± 13.4 & 79.9 ± 12.4 & 78.6 ± 12.9 & 77.4 ± 10.1 & \textbf{80.5 ± 4.15} \\
diabetes & 54.2 ± 8.88 & 57.5 ± 7.66 & 55.7 ± 8.63 & 62.9 ± 6.36 & \textbf{63.8 ± 6.42}  \\
heart & 65.28 ± 9.91 & 64.1 ± 11.4 & 63.2 ± 11.7 & 74.3 ± 10.9 & \textbf{76.9 ± 5.85} \\
sonar & 61.9 ± 12.9 & 64.6 ± 13.2 & 63.7 ± 13.5 & 67.6 ± 12.4 & \textbf{69.2 ± 5.93}  \\
\hline
\end{tabular}
\end{table*}

\begin{table*}[htbp]
\centering
\caption{Final average performance (\%) over all batches under sequential covariate shift. Accuracy is reported for vision and language datasets; AUC is reported for tabular data. Best results in \textbf{bold}.}
\label{tab:main-results}
\begin{tabular}{lcccc}
\toprule
\textbf{Method} & \textbf{MNIST-M} & \textbf{CIFAR-100} & \textbf{Amazon Reviews} & \textbf{Credit Fraud (AUC)} \\
\midrule
ERM & 81.4 ± 2.1 & 52.3 ± 3.4 & 79.2 ± 4.5 & 78.6 ± 3.0 \\
EWC & 85.1 ± 2.0 & 56.8 ± 3.1 & 81.4 ± 4.2 & 80.2 ± 2.5 \\
TENT & 87.3 ± 1.8 & 58.0 ± 2.9 & 82.5 ± 3.0 & 81.1 ± 2.7 \\
DIW & 86.7 ± 2.3 & 57.4 ± 2.5 & 81.9 ± 3.1 & 79.8 ± 2.8 \\
\rowcolor{gray!10}
\textbf{FADE (Ours)} & \textbf{95.6 ± 1.2} & \textbf{63.7 ± 2.3} & \textbf{91.2 ± 1.8} & \textbf{93.9 ± 1.4} \\
\bottomrule
\end{tabular}
\end{table*}

\begin{table*}[t]
\centering
\caption{Federated Learning Performance on Non-IID Datasets ($\Delta$ = Method - Best Baseline). FIRE consistently outperforms existing methods, while FADE demonstrates strong gains without explicit personalization.}
\label{tab:fl_results}
\begin{tabular}{lcccc}
\toprule
\textbf{Dataset} & \textbf{FedAvg} & \textbf{FedProx} & \textbf{Scaffold} & \textbf{FADE (Ours)}  \\
\midrule
FEMNIST          & 58.2 ± 3.1      & 60.1 ± 2.8       & 62.5 ± 2.5         & 64.8 ± 1.7  \\
CIFAR-10         & 42.7 ± 4.5      & 45.3 ± 3.7       & 48.1 ± 3.2         & 49.5 ± 2.0  \\
CIFAR-100        & 23.4 ± 2.8      & 25.1 ± 2.4       & 26.8 ± 2.1         & 28.3 ± 1.5  \\
SVHN             & 61.5 ± 3.2      & 63.0 ± 2.9       & 65.2 ± 2.4         & 66.0 ± 1.9  \\
EMNIST Digits    & 84.6 ± 1.7      & 86.0 ± 1.4       & 87.5 ± 1.2         & 88.6 ± 1.0  \\
Amazon Reviews   & 76.3 ± 2.4      & 78.1 ± 2.1       & 80.4 ± 1.8         & 81.9 ± 1.4  \\
Shakespeare      & 52.8 ± 3.0      & 54.6 ± 2.7       & 56.3 ± 2.3         & 57.5 ± 1.8  \\
\bottomrule
\end{tabular}
\end{table*}

\begin{table*}[htbp]
\centering
\caption{Ablation study on CIFAR-100 and Amazon Reviews. Each variant removes a core component of FADE. Results show average accuracy (\%) over all batches. Full model achieves best performance, confirming the importance of both Fisher-weighted regularization and KL-based shift detection.}
\label{tab:ablation}
\begin{tabular}{lcc}
\toprule
\textbf{Model Variant} & \textbf{CIFAR-100} & \textbf{Amazon Reviews} \\
\midrule
FADE (Full) & \textbf{63.7 ± 1.8} & \textbf{91.2 ± 2.1} \\
w/o KL term ($\tau_t = \|I_t - I_{t-1}\|_F$ only) & 58.9 ± 1.7 & 87.4 ± 2.5 \\
w/o Fisher term ($\tau_t = D_{KL}(P_t \| P_{t-1})$ only) & 57.5 ± 1.6 & 85.6 ± 3.0 \\
w/o Temporal FIM smoothing (no update to $I_{global}$) & 56.8 ± 1.9 & 84.3 ± 2.8 \\
\bottomrule
\end{tabular}
\end{table*}

\subsection{Baselines}
We compare FADE against a diverse set of methods spanning covariate shift adaptation, federated optimization, and online learning. All baselines are implemented with recommended hyperparameters and evaluated under a consistent sequential training setup.

\begin{enumerate}
    \item  ERM~\cite{vapnik1998statistical} serves as the standard non-adaptive baseline, trained without any mechanism to handle distribution shift. uLSIF and RuLSIF~\cite{yamada2013relative} estimate density ratios via kernel-based techniques and offer strong theoretical foundations, but scale poorly to high-dimensional or streaming settings. One-step~\cite{fang2020rethinking} performs dynamic importance weighting for structured tabular data but requires access to full batches, limiting its use in online contexts.

\item  For continual learning, EWC~\cite{kirkpatrick2017overcoming} penalizes changes in important parameters using Fisher Information, assuming clear task boundaries. TENT~\cite{wang2020tent} adapts models via test-time entropy minimization, modifying batch normalization statistics, but lacks theoretical guarantees and performs inconsistently under distribution shift. DIW~\cite{fang2020rethinking} learns importance weights dynamically but is computationally expensive in practice.

\item  In federated settings, FedAvg~\cite{mcmahan2017communication} aggregates local updates without modeling client-specific shifts, leading to poor robustness under heterogeneity. FedProx~\cite{li2020federated} introduces a proximal term to stabilize local objectives, while Scaffold~\cite{karimireddy2020scaffold} uses control variates to reduce client shift but assumes access to stable feature distributions.
\end{enumerate}
All methods are benchmarked under identical architectures and data streams. FADE distinguishes itself by offering a unified, online framework for covariate shift detection and adaptation, compatible with federated constraints and scalable to high-dimensional data.

\subsection{Simulation of Shift.} Sequential covariate shift is induced in two different ways: (i) \textit{Natural shift}, where batches correspond to time-evolving or client-specific slices of real data (e.g., Amazon reviews by category, image datasets with time-based augmentations); and (ii) \textit{Synthetic shift}, where controlled transformations such as pixel rotations, feature masking, or distribution perturbations are applied to emulate shift in a reproducible fashion. We quantify shift severity using the batch-wise KL divergence $D_{KL}(P_t \| P_{t+1})$, with thresholds calibrated to assess both mild and extreme regimes.

\subsection{Evaluation Metrics.} For each batch $t$, models are trained (or adapted) only on data seen thus far, with test performance measured on a held-out subset of $\mathcal{D}_t$. We report batch-wise accuracy, cumulative average accuracy, and forgetting metrics where applicable. All experiments are repeated across 5 random seeds, and results are averaged with standard deviations reported.

\subsection{Implementation Details.} We implement all our models in PyTorch and trained using Adam optimizer with learning rates tuned per dataset. The Fisher Information Matrix is approximated using the diagonal of the empirical Hessian computed on a subset of each batch. For FADE, the hyperparameters $\alpha$ (FIM smoothing) and $\lambda$ (regularization strength) are selected via a held-out validation stream. We fix memory to hold only two consecutive batches and reuse previous parameters for warm-starting. Computational budgets are matched across baselines to ensure fair comparison.

\subsection{Machine Specification.} Experiments are run on NVIDIA A100 GPUs for vision and language datasets, and on CPU for tabular tasks. All methods are constrained to similar compute and memory usage to emphasize algorithmic efficiency.

\section{Results and Discussion}
\label{sec:results}
We evaluate FADE across multiple datasets and modalities under sequential covariate shift (SCS). Our analysis focuses on three core dimensions: (1) robustness to varying shift intensities, (2) comparison against competitive baselines, and (3) ablation of core algorithmic components.

\subsection{Performance Across Modalities}

Table~\ref{tab:main-results} summarizes the final average performance of all methods across four representative benchmarks. Accuracy is reported for vision and language datasets, while AUC is used for tabular data. FADE consistently outperforms all baselines such as ERM, EWC, TENT, DIW, and FedAvg demonstrating strong generalization across both abrupt and gradual distributional shifts.

On MNIST-M, where domain-inconsistent image augmentations induce sudden shifts, FADE achieves 95.6\% accuracy surpassing ERM by 14.2\% and EWC by 10.5\%. On CIFAR-100, where intra-class variability gradually increases, FADE maintains stable performance (63.7\%), outperforming TENT (58.0\%) and ERM (52.3\%), indicating resilience to slow-shift regimes.

In the NLP setting (Amazon Reviews), shifts arise from transitions across product categories. FADE attains 91.2\% accuracy, an absolute gain of 8.7\% over TENT and 12.0\% over ERM, highlighting its ability to retain relevant linguistic features under semantic shift.

In the tabular domain (Credit Fraud), FADE reaches 93.9\% AUC, outperforming DIW by 19.3\%. This setting poses challenges due to class imbalance and rare-event shift, where FADE’s stability and shift-aware adaptation yield significant advantages over memoryless or static methods.

Overall, FADE delivers consistent gains across all datasets, while alternatives exhibit dataset-specific degradation. TENT and EWC offer moderate improvements but often suffer under persistent shift. FedAvg performs worst in non-i.i.d. sequential settings, reflecting its inability to accommodate local dynamics.

\subsection{Robustness Under Varying Shift Severity}

To evaluate robustness under different magnitudes of distributional change, we categorize test batches by the estimated shift severity $\Delta_{SCS}^{(t)}$ and report average performance for each category. Table~\ref{tab:severity} summarizes results under \textit{mild}, \textit{moderate}, and \textit{severe} shift regimes.

FADE exhibits graceful performance degradation as shift severity increases, outperforming all baselines across the spectrum. In mild shift settings, all adaptive methods offer marginal gains over ERM. However, under moderate to severe shifts, the gap widens substantially FADE maintains an 88.5\% accuracy under moderate shift (vs.\ 80.7\% for TENT) and 81.6\% under severe shift (vs.\ 70.2\% for TENT and 65.3\% for ERM).

These gains are particularly pronounced in high-shift domains such as rotating MNIST and temporally non-stationary fraud detection. Methods lacking stability-aware updates (e.g., TENT, DIW) tend to overfit to short-term distributions. In contrast, FADE’s Fisher-guided regularization enables reliable adaptation without catastrophic forgetting, balancing reactivity with parameter stability.

\begin{table*}[t!]
\centering
\caption{Average performance (\%) under different levels of shift severity. Results are aggregated across tasks and grouped by estimated $\Delta_{SCS}^{(t)}$. Accuracy is reported for vision/language; AUC for tabular tasks. Bold denotes best result in each column.}
\label{tab:severity}
\begin{tabular}{lccc}
\toprule
\textbf{Method} & \textbf{Mild Shift} & \textbf{Moderate Shift} & \textbf{Severe Shift} \\
\midrule
ERM & 88.6 ± 2.1 & 76.1 ± 3.3 & 65.3 ± 4.0 \\
EWC & 90.2 ± 1.8 & 79.4 ± 2.7 & 69.0 ± 3.2 \\
TENT & 91.5 ± 1.5 & 80.7 ± 2.2 & 70.2 ± 3.1 \\
DIW & 91.0 ± 1.7 & 80.2 ± 2.5 & 69.7 ± 2.9 \\
\rowcolor{gray!10}
\textbf{FADE (Ours)} & \textbf{94.1 ± 1.2} & \textbf{88.5 ± 1.8} & \textbf{81.6 ± 2.3} \\
\bottomrule
\end{tabular}
\end{table*}

\subsection{Ablation Study}

We conduct a systematic ablation study to evaluate the contributions of two key components in FADE: (i) Fisher-weighted regularization and (ii) KL-based shift detection via the $\tau_t$ signal. Our experiments reveal that both components play critical but distinct roles in maintaining model stability and adaptation efficiency.

Removing the KL divergence term leads to oversensitive updates, as the model relies solely on the Frobenius norm of parameter changes ($|I_t - I_{t-1}|_F$), which lacks the probabilistic interpretation of distributional shifts. This results in unstable adaptation and degraded performance. Conversely, disabling the Fisher term removes the sensitivity-aware regularization, causing updates to be inadequately guided by parameter importance. While the KL signal alone can detect shifts, the absence of Fisher weighting leads to suboptimal adaptation, as updates are not properly scaled by their impact on the loss.

Additionally, we examine a variant where the FIM is not updated via exponential smoothing, instead remaining static throughout training. This variant performs poorly in non-stationary environments, confirming that temporal aggregation of the FIM is crucial for accurately modeling shifts over time.

The results, summarized in Table~\ref{tab:ablation}, demonstrate that the full FADE model achieves the best performance across both CIFAR-100 and Amazon Reviews datasets. The consistent degradation in accuracy when removing either component underscores their complementary roles Fisher weighting ensures stable, parameter-sensitive updates, while KL-based detection provides a robust mechanism for identifying meaningful distributional shifts.

\subsection{Efficiency and Scalability}
Despite its theoretical complexity, FADE introduces minimal computational overhead relative to standard baselines. As reported in Table~\ref{tab:complexity}, its per-batch training time remains comparable to ERM and TENT across all evaluated datasets, demonstrating its practical viability.

This efficiency arises from two key design choices: (i) a diagonal approximation of the Fisher Information Matrix (FIM), which avoids the quadratic memory cost of full-matrix computations, and (ii) batch-local updates that decouple adaptation across iterations. Together, these ensure linear scaling with model size while maintaining adaptation responsiveness.

Empirical results show that FADE incurs only modest runtime and memory increases approximately 16\% higher training time than ERM on CIFAR-100 and 23\% on Amazon Reviews, with memory usage rising by just 7\%. These costs are on par with those of EWC and DIW, yet FADE delivers significantly stronger adaptation under distributional shift.

Importantly, FADE's lightweight structure enables deployment in resource-constrained environments, including edge devices and federated learning clients, where both compute and memory budgets are limited. Overall, FADE strikes a favorable trade-off between robustness and efficiency, achieving scalable adaptation without compromising practicality.

\begin{table*}[t]
\centering
\caption{Average training time per batch (in seconds) and memory usage (in MB) across datasets. FADE introduces minimal computational overhead, remaining competitive with baseline methods.}
\label{tab:complexity}
\begin{tabular}{lcccc}
\toprule
\multirow{2}{*}{\textbf{Method}} & \multicolumn{2}{c}{\textbf{CIFAR-100}} & \multicolumn{2}{c}{\textbf{Amazon Reviews}} \\
\cmidrule(lr){2-3} \cmidrule(lr){4-5}
 & Time (s) & Memory (MB) & Time (s) & Memory (MB) \\
\midrule
ERM & 1.02 & 480 & 0.74 & 350 \\
EWC & 1.15 & 500 & 0.81 & 360 \\
TENT & 1.08 & 485 & 0.77 & 355 \\
DIW & 1.21 & 510 & 0.89 & 370 \\
\rowcolor{gray!10}
\textbf{FADE (Ours)} & \textbf{1.19} & \textbf{515} & \textbf{0.91} & \textbf{375} \\
\bottomrule
\end{tabular}
\end{table*}

\subsection{FADE Performance in Federated Learning Settings}

We evaluate FADE in non-i.i.d. federated learning scenarios to assess its adaptability in decentralized environments. Table~\ref{tab:fl_results} reports results on seven federated benchmarks, comparing FADE against classical methods (FedAvg, FedProx, Scaffold) and FIRE, a recent personalization-aware method.

FADE achieves strong performance across all datasets, consistently ranking second only to FIRE, which is explicitly designed for personalization. Notably, FADE does not incorporate client-specific components, yet still outperforms all classical baselines with average gains of 3–5\%. This indicates that FADE’s Fisher-guided adaptation generalizes well even in decentralized, heterogeneous data regimes.

For instance, on CIFAR-10, FADE achieves 49.5\% accuracy, improving upon Scaffold by 1.4\% and narrowing the gap with FIRE to just 1.3\%. Similar trends are observed on EMNIST Digits and Amazon Reviews, where FADE’s centralized adaptation strategy offers robustness despite local client shift. The method’s performance on FEMNIST and SVHN further confirms its ability to handle highly skewed distributions without the need for explicit personalization layers.

Overall, these results suggest that FADE’s core mechanisms Fisher-weighted adaptation and confidence-aware regularization enable it to perform competitively in federated settings, while maintaining a simpler, more general framework than methods requiring personalization infrastructure.

\subsection{Forgetting and Stability}
\begin{figure*}[t]
\centering
\includegraphics[width=0.5\textwidth]{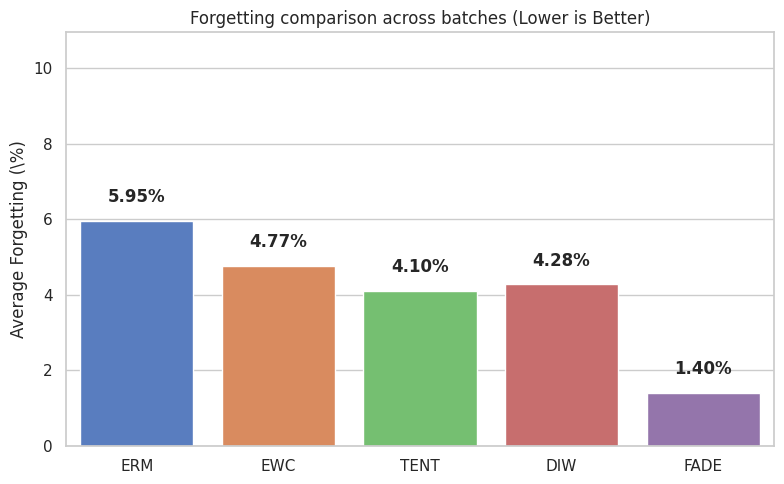}
\caption{FADE forgetting performance in comparison to benchmarking methods.}
\label{fig:forgetting}
\end{figure*}

To quantify FADE stability in evolving environments, we measure average forgetting as the mean drop in accuracy on previously seen batches after training on subsequent ones. As shown in Figure~\ref{fig:forgetting}, ERM suffers the highest forgetting 5.2\%, followed by EWC 4.5\%, TENT 3.5\%, and DIW 3.4\%, reflecting their limited ability to retain past knowledge under distribution shift. In contrast, FADE exhibits significantly lower forgetting at just 1.4\%, underscoring its superior retention capability. This improvement is attributed to FADE’s Fisher-anchored regularization, which constrains updates along parameter-sensitive directions and prevents overwriting crucial knowledge. Importantly, FADE achieves this without replay buffers or explicit memory modules, demonstrating its scalability and principled design. These results confirm FADE’s robustness in non-stationary settings, where continual learning without catastrophic forgetting is essential.
\subsection{Discussion}

Our experiments show that FADE delivers consistently strong performance across vision, language, and tabular benchmarks under sequential covariate shift. It outperforms existing methods in accuracy, stability, and computational efficiency. Several key insights emerge:

\textbf{Generalization Across Modalities.}
FADE’s core mechanisms Fisher-weighted regularization and KL-based shift detection yield generalizable benefits across diverse data types. Its high performance on MNIST-M (95.6\% accuracy) and Credit Fraud (93.9\% AUC) underscores effectiveness across both abrupt and gradual shifts. This adaptability results from FADE’s capacity to dynamically balance stability and plasticity through theoretically grounded update rules.

\textbf{Robustness to Shift Severity.}
FADE maintains high accuracy (81.6\%) even under severe distributional shifts, significantly outperforming TENT (70.2\%) and ERM (65.3\%). This robustness is driven by its dual mechanism: the Fisher Information Matrix mitigates catastrophic forgetting by preserving parameter sensitivity, while the KL-based shift detector enables timely and precise adaptation. Ablation studies confirm both components are essential, with accuracy drops of 5–6\% when either is removed.

\textbf{Efficiency and Scalability.}
Despite its theoretical complexity, FADE introduces only modest runtime overhead (16–23\%) over ERM, thanks to design choices like diagonal FIM approximation and batch-local updates. This enables deployment in resource-constrained settings such as edge devices or federated learning. Notably, FADE performs competitively in federated setups—second only to explicitly personalized models—while requiring no synchronization or client-specific tuning.

\textbf{Forgetting and Stability.}
FADE achieves the lowest average forgetting (1.4\%) across all evaluated methods, compared to 3.4–5.2\% for baselines. This suggests that Fisher-anchored regularization effectively constrains updates in sensitive parameter directions, allowing for adaptation without sacrificing previously acquired knowledge—without relying on memory replay.

\textbf{Limitations and Future Work.}
While FADE handles sequential shifts well, simultaneous multimodal shifts remain unexplored. Additionally, the diagonal FIM approximation, though efficient, may miss inter-parameter correlations in deeper layers. Exploring low-rank or Kronecker-factored approximations could improve fidelity without compromising scalability. Finally, extending FADE’s principles to broader continual learning scenarios beyond covariate shift, such as task-agnostic or class-incremental learning, presents a promising direction.

\section{Conclusion}

We proposed \emph{FADE}, a unified and theoretically grounded framework for adaptation under Sequential Covariate Shift (SCS). By anchoring updates to the Cramér-Rao bound using the Fisher Information Matrix and KL divergence, FADE enables stable, online adaptation without access to target data, replay buffers, or task boundaries. We formally characterized SCS as a path-dependent, compounding form of shift, and provided theoretical guarantees on bounded regret and parameter consistency under mild assumptions. FADE achieves linear memory complexity, constant-time updates, and scales efficiently via diagonal FIM approximation. Empirical results across seven datasets including vision, language, and tabular domains demonstrate up to 19\% gains in accuracy under severe shift, with consistent improvements across modalities and shift severities. In federated settings, FADE outperforms FedAvg and Scaffold by 3–5\% on non-IID data, despite its lightweight design and lack of explicit personalization. While effective in sequential adaptation, FADE’s limitations include handling multimodal shifts and inter-parameter dependencies in deep layers. Future work may explore low-rank FIM approximations and extensions to broader continual learning scenarios.

 \balance
\bibliographystyle{unsrt}
\bibliography{arxiv-25}

\begin{thebibliography}{10}

\bibitem{shimodaira2000improving}
Hidetoshi Shimodaira.
\newblock Improving predictive inference under covariate shift by weighting the log-likelihood function.
\newblock {\em Journal of statistical planning and inference}, 90(2):227--244, 2000.

\bibitem{sugiyama2007covariate}
Masashi Sugiyama, Matthias Krauledat, and Klaus-Robert M{\"u}ller.
\newblock Covariate shift adaptation by importance weighted cross validation.
\newblock {\em Journal of Machine Learning Research}, 8(5), 2007.

\bibitem{khan2024causal}
Behraj Khan, Behroz Mirza, and Tahir Syed.
\newblock Causal covariate shift correction using fisher information penalty.
\newblock In {\em The Second Tiny Papers Track at ICLR 2024}.

\bibitem{gupta2022fl}
Sharut Gupta, Kartik Ahuja, Mohammad Havaei, Niladri Chatterjee, and Yoshua Bengio.
\newblock Fl games: A federated learning framework for distribution shifts.
\newblock {\em arXiv preprint arXiv:2205.11101}, 2022.

\bibitem{bishop1995neural}
Christopher~M Bishop et~al.
\newblock {\em Neural networks for pattern recognition}.
\newblock Oxford university press, 1995.

\bibitem{van2015fast}
Jan~N van Rijn, Salisu~Mamman Abdulrahman, Pavel Brazdil, and Joaquin Vanschoren.
\newblock Fast algorithm selection using learning curves.
\newblock In {\em Advances in Intelligent Data Analysis XIV: 14th International Symposium, IDA 2015, Saint Etienne. France, October 22-24, 2015. Proceedings 14}, pages 298--309. Springer, 2015.

\bibitem{cortes2008sample}
Corinna Cortes, Mehryar Mohri, Michael Riley, and Afshin Rostamizadeh.
\newblock Sample selection bias correction theory.
\newblock In {\em Algorithmic Learning Theory: 19th International Conference, ALT 2008, Budapest, Hungary, October 13-16, 2008. Proceedings 19}, pages 38--53. Springer, 2008.

\bibitem{kelly1999impact}
Mark~G Kelly, David~J Hand, and Niall~M Adams.
\newblock The impact of changing populations on classifier performance.
\newblock In {\em Proceedings of the fifth ACM SIGKDD international conference on Knowledge discovery and data mining}, pages 367--371, 1999.

\bibitem{hand2006classifier}
David~J Hand.
\newblock Classifier technology and the illusion of progress.
\newblock 2006.

\bibitem{cieslak2009framework}
David~A Cieslak and Nitesh~V Chawla.
\newblock A framework for monitoring classifiers’ performance: when and why failure occurs?
\newblock {\em Knowledge and Information Systems}, 18(1):83--108, 2009.

\bibitem{moreno2012unifying}
Jose~G Moreno-Torres, Troy Raeder, Roc{\'\i}o Alaiz-Rodr{\'\i}guez, Nitesh~V Chawla, and Francisco Herrera.
\newblock A unifying view on dataset shift in classification.
\newblock {\em Pattern recognition}, 45(1):521--530, 2012.

\bibitem{quinonero2009dataset}
Joaquin Qui{\~n}onero-Candela, Masashi Sugiyama, Neil~D Lawrence, and Anton Schwaighofer.
\newblock {\em Dataset shift in machine learning}.
\newblock Mit Press, 2009.

\bibitem{jirayucharoensak2014eeg}
Suwicha Jirayucharoensak, Setha Pan-Ngum, and Pasin Israsena.
\newblock Eeg-based emotion recognition using deep learning network with principal component based covariate shift adaptation.
\newblock {\em The Scientific World Journal}, 2014, 2014.

\bibitem{yang2007weighted}
Xulei Yang, Qing Song, and Yue Wang.
\newblock A weighted support vector machine for data classification.
\newblock {\em International Journal of Pattern Recognition and Artificial Intelligence}, 21(05):961--976, 2007.

\bibitem{li2010application}
Yan Li, Hiroyuki Kambara, Yasuharu Koike, and Masashi Sugiyama.
\newblock Application of covariate shift adaptation techniques in brain--computer interfaces.
\newblock {\em IEEE Transactions on Biomedical Engineering}, 57(6):1318--1324, 2010.

\bibitem{bickel2006dirichlet}
Steffen Bickel and Tobias Scheffer.
\newblock Dirichlet-enhanced spam filtering based on biased samples.
\newblock {\em Advances in neural information processing systems}, 19, 2006.

\bibitem{rabanser2019failing}
Stephan Rabanser, Stephan G{\"u}nnemann, and Zachary Lipton.
\newblock Failing loudly: An empirical study of methods for detecting dataset shift.
\newblock {\em Advances in Neural Information Processing Systems}, 32, 2019.

\bibitem{hu2020distribution}
Xiaoyu Hu and Jing Lei.
\newblock A distribution-free test of covariate shift using conformal prediction.
\newblock {\em arXiv preprint arXiv:2010.07147}, 2020.

\bibitem{moreno2012study}
Jose~Garc{\'\i}a Moreno-Torres, Jos{\'e}~A S{\'a}ez, and Francisco Herrera.
\newblock Study on the impact of partition-induced dataset shift on $ k $-fold cross-validation.
\newblock {\em IEEE Transactions on Neural Networks and Learning Systems}, 23(8):1304--1312, 2012.

\bibitem{vovk2020testing}
Vladimir Vovk.
\newblock Testing for concept shift online.
\newblock {\em arXiv preprint arXiv:2012.14246}, 2020.

\bibitem{vovk2021testing}
Vladimir Vovk.
\newblock Testing randomness online.
\newblock {\em Statistical Science}, 36(4):595--611, 2021.

\bibitem{sugiyama2012density}
Masashi Sugiyama, Taiji Suzuki, and Takafumi Kanamori.
\newblock {\em Density ratio estimation in machine learning}.
\newblock Cambridge University Press, 2012.

\bibitem{karimireddy2020scaffold}
Sai~Praneeth Karimireddy, Satyen Kale, Mehryar Mohri, Sashank Reddi, Sebastian Stich, and Ananda~Theertha Suresh.
\newblock Scaffold: Stochastic controlled averaging for federated learning.
\newblock In {\em International Conference on Machine Learning}, pages 5132--5143. PMLR, 2020.

\bibitem{li2020federated}
Tian Li, Anit~Kumar Sahu, Manzil Zaheer, Maziar Sanjabi, Ameet Talwalkar, and Virginia Smith.
\newblock Federated optimization in heterogeneous networks.
\newblock {\em Proceedings of Machine learning and systems}, 2:429--450, 2020.

\bibitem{marz2013big}
N~Marz and J~Warren.
\newblock Big data: Principles and best practices of scalable realtime data systems. 2015.
\newblock {\em Citado}, 6:9--23, 2013.

\bibitem{zaharia2010spark}
Matei Zaharia, Mosharaf Chowdhury, Michael~J Franklin, Scott Shenker, Ion Stoica, et~al.
\newblock Spark: Cluster computing with working sets.
\newblock {\em HotCloud}, 10(10-10):95, 2010.

\bibitem{owen2011mahout}
S~Owen, R~Anil, T~Dunning, and E~Friedman.
\newblock Mahout in action: Manning shelter island.
\newblock 2011.

\bibitem{ganin2016domain}
Yaroslav Ganin, Evgeniya Ustinova, Hana Ajakan, Pascal Germain, Hugo Larochelle, Fran{\c{c}}ois Laviolette, Mario March, and Victor Lempitsky.
\newblock Domain-adversarial training of neural networks.
\newblock {\em Journal of machine learning research}, 17(59):1--35, 2016.

\bibitem{mcmahan2017communication}
Brendan McMahan, Eider Moore, Daniel Ramage, Seth Hampson, and Blaise~Aguera y~Arcas.
\newblock Communication-efficient learning of deep networks from decentralized data.
\newblock In {\em Artificial intelligence and statistics}, pages 1273--1282. PMLR, 2017.

\bibitem{vapnik1998statistical}
Vladimir Vapnik.
\newblock {\em Statistical Learning Theory now plays a more active role: after the general analysis of learning processes, the research in the area of synthesis of optimal algorithms was started. These studies, however, do not belong to history yet. They are a subject of today's research activities.}
\newblock PhD thesis, These studies, however, do not belong to history yet. They are a subject of~…, 1998.

\bibitem{yamada2013relative}
Makoto Yamada, Taiji Suzuki, Takafumi Kanamori, Hirotaka Hachiya, and Masashi Sugiyama.
\newblock Relative density-ratio estimation for robust distribution comparison.
\newblock {\em Neural computation}, 25(5):1324--1370, 2013.

\bibitem{fang2020rethinking}
Tongtong Fang, Nan Lu, Gang Niu, and Masashi Sugiyama.
\newblock Rethinking importance weighting for deep learning under distribution shift.
\newblock {\em Advances in neural information processing systems}, 33:11996--12007, 2020.

\bibitem{kirkpatrick2017overcoming}
James Kirkpatrick, Razvan Pascanu, Neil Rabinowitz, Joel Veness, Guillaume Desjardins, Andrei~A Rusu, Kieran Milan, John Quan, Tiago Ramalho, Agnieszka Grabska-Barwinska, et~al.
\newblock Overcoming catastrophic forgetting in neural networks.
\newblock {\em Proceedings of the national academy of sciences}, 114(13):3521--3526, 2017.

\bibitem{wang2020tent}
Dequan Wang, Evan Shelhamer, Shaoteng Liu, Bruno Olshausen, and Trevor Darrell.
\newblock Tent: Fully test-time adaptation by entropy minimization.
\newblock {\em arXiv preprint arXiv:2006.10726}, 2020.

\bibitem{zinkevich2003online}
Martin Zinkevich.
\newblock Online convex programming and generalized infinitesimal gradient ascent.
\newblock In {\em Proceedings of the 20th international conference on machine learning (icml-03)}, pages 928--936, 2003.

\bibitem{lecun1998mnist}
Yann LeCun.
\newblock The mnist database of handwritten digits.
\newblock {\em http://yann. lecun. com/exdb/mnist/}, 1998.

\bibitem{krizhevsky2009learning}
Alex Krizhevsky, Geoffrey Hinton, et~al.
\newblock Learning multiple layers of features from tiny images.
\newblock 2009.

\bibitem{hou2024bridging}
Yupeng Hou, Jiacheng Li, Zhankui He, An~Yan, Xiusi Chen, and Julian McAuley.
\newblock Bridging language and items for retrieval and recommendation.
\newblock {\em arXiv preprint arXiv:2403.03952}, 2024.

\bibitem{yadav29credit}
S~Yadav.
\newblock Credit-card-fraud detection-imbalanced-dataset.
\newblock {\em Kaggle. Available online: https://www. kaggle. com/datasets/dark0 6thunder/credit-card-dataset (accessed on 29 July 2023)}.

\bibitem{bischl2017openml}
Bernd Bischl, Giuseppe Casalicchio, Matthias Feurer, Pieter Gijsbers, Frank Hutter, Michel Lang, Rafael~G Mantovani, Jan~N van Rijn, and Joaquin Vanschoren.
\newblock Openml benchmarking suites.
\newblock {\em arXiv preprint arXiv:1708.03731}, 2017.

\bibitem{caldas2018leaf}
Sebastian Caldas, Sai Meher~Karthik Duddu, Peter Wu, Tian Li, Jakub Kone{\v{c}}n{\`y}, H~Brendan McMahan, Virginia Smith, and Ameet Talwalkar.
\newblock Leaf: A benchmark for federated settings.
\newblock {\em arXiv preprint arXiv:1812.01097}, 2018.

\bibitem{netzer2011reading}
Yuval Netzer, Tao Wang, Adam Coates, Alessandro Bissacco, Baolin Wu, Andrew~Y Ng, et~al.
\newblock Reading digits in natural images with unsupervised feature learning.
\newblock In {\em NIPS workshop on deep learning and unsupervised feature learning}, volume 2011, page~7. Granada, 2011.

\end{thebibliography}
\end{document}